\pdfoutput=1
\def\ARXIVVERSION{}
\documentclass[letterpaper, 10 pt, conference]{ieeeconf}
\IEEEoverridecommandlockouts
\usepackage{amsmath,amssymb}
\usepackage{graphicx}
\usepackage{booktabs}
\usepackage{cite}
\usepackage{tikz}
\usetikzlibrary{arrows.meta,positioning,calc}
\usepackage{url}
\urldef{\repo}\url{https://github.com/Yunfei-Bi8/Delay-tolerant-agv-formation}

\newif\ifarxiv
\ifdefined\ARXIVVERSION\arxivtrue\else\arxivfalse\fi
\newcommand{\proofsin}{\ifarxiv the Appendix\else\cite{bi2026arxiv}\fi}
\newcommand{\corref}{\ifarxiv Corollary~\ref{cor:iss} in the Appendix\else\cite[Cor.~1]{bi2026arxiv}\fi}

\newtheorem{assumption}{Assumption}
\newtheorem{lemma}{Lemma}
\newtheorem{theorem}{Theorem}
\newtheorem{corollary}{Corollary}
\newtheorem{proposition}{Proposition}
\newtheorem{remark}{Remark}
\renewenvironment{proof}[1][Proof]{\par\noindent\hspace{1em}{\itshape #1:}\ }{\hspace*{\fill}~$\blacksquare$\par}

\newcommand{\R}{\mathbb{R}}
\newcommand{\eps}{\varepsilon}
\newcommand{\vt}{\vartheta}
\newcommand{\xip}{\xi^{\perp}}
\newcommand{\sinc}{\operatorname{sinc}}
\newcommand{\vh}{\hat v}
\newcommand{\wh}{\hat\omega}
\newcommand{\thh}{\hat\theta}
\newcommand{\vvh}{\hat{\mathbf v}}
\newcommand{\avh}{\hat{\mathbf a}}
\newcommand{\ee}{\mathrm{e}}

\title{\LARGE \bf Delay-Tolerant Formation Tracking for Multi-AGV Systems\\ via Reference-Shift Coupling}

\author{Yunfei Bi%
\thanks{Y.~Bi is with the Technical University of Munich, Munich, Germany. {\tt\small yunfei.bi@tum.de}}%
\thanks{Code, data and the scripts for all figures and tables: \repo.}%
}

\begin{document}

\maketitle
\thispagestyle{empty}
\pagestyle{empty}

%%%%%%%%%%%%%%%%%%%%%%%%%%%%%%%%%%%%%%%%%%%%%%%%%%%%%%%%%%%%%%%%%%%%%%%%%%%%%%%%
\begin{abstract}
Automated guided vehicles (AGVs) that move in formation must keep their relative positions while each vehicle tracks
its own reference, over wireless links with time-varying, link-dependent delays. In the proposed cascade controller for
unicycle-type AGVs, delayed neighbor data drive only a saturated second-order filter that shifts each vehicle's
reference, and an unmodified gain-scheduled tracking law, whose Lyapunov function is therefore independent of the
network, follows the shifted reference. The coupling cannot destroy tracking: for every bounded delay on directed,
switching graphs, the errors converge exponentially on bounded sets, with delay-independent constants.
On undirected graphs, it reduces the disagreement caused by persistent local errors below a design bound, by a factor
that grows with the coupling gain and the algebraic connectivity. A direct coupling of the delayed errors, in contrast,
is stable for every delay only below a gain threshold, beyond which it lost stability in simulation as predicted. In
simulations of three AGVs with latencies of tens of milliseconds, the cascade reduced the formation error after traction
losses and safety stops at a cost in tracking accuracy and never failed; the tuned direct coupling did better after
traction losses, about half of its benefit stemming from its higher gain. With latencies of 2\,s or more, the cascade
increased the formation error unless outdated samples were discarded.
\end{abstract}

%%%%%%%%%%%%%%%%%%%%%%%%%%%%%%%%%%%%%%%%%%%%%%%%%%%%%%%%%%%%%%%%%%%%%%%%%%%%%%%%
\section{INTRODUCTION}

Teams of automated guided vehicles (AGVs) move in formation, for instance to carry a large load on compliant fixtures
\cite{peng2023}. We consider vehicles that are not rigidly connected: each one follows its planned trajectory, and the
formation is kept by control over wireless links on which the latency varies from packet to packet and packets are lost
in bursts. Since a receiver can only use the newest sample it has received, every link introduces its own time-varying
delay, which drops at each packet arrival.

In \emph{direct coupling}, the tracking errors of the neighbors enter the tracking law of each vehicle
\cite{kostic2010,sadowska2011,sadowska2012}; over a network, the delay then acts inside the nonlinear vehicle loop, and
the analysis guarantees stability up to an admissible delay bound \cite{alvarez2011}. For nonholonomic vehicles with
time-varying delays, consensus and rendezvous have been solved mostly at the force level and on undirected graphs
\cite{nuno2020,maghenem2021,maghenem2021lcss,romero2024,paredes2025}, and leader--follower formations in
\cite{loria2021}; in \cite{nuno2022}, only controller states cross the delayed network. In a \emph{cascade}, the network
acts only on a reference generator, and each vehicle loop can be analyzed as if there were no delay. This structure has
been established for fully actuated Euler--Lagrange systems \cite{abdessameud2014,wang2014,peng2023}, for underactuated
thrust-propelled vehicles that converge to a stationary formation \cite{abdessameud2011,abdessameud2015}, and for
coordinated path following, where a scalar path parameter is exchanged \cite{ghabcheloo2009,li2012}; in
\cite{maghenem2018}, a cascade structure is used to analyze formation tracking without delays. For nonholonomic
vehicles, separating consensus from stabilization has been considered ineffective, because a reference produced by a
consensus protocol need not be feasible \cite{maghenem2021lcss}.

For references whose speed is bounded away from zero, this paper shows that saturating the reference shift removes
this obstacle. Building on \cite{peng2023}, it develops a delay-tolerant cascade for kinematic unicycles that track
individual time-parameterized trajectories, the level at which industrial AGVs are commanded: a second-order filter
alone receives the delayed data of the neighbors and shifts the vehicle's own reference, and an unmodified
gain-scheduled tracking law \cite{samson1993,deluca2001} follows the shifted reference. The contributions are:
\begin{enumerate}
\item Saturations keep the shifted reference a unicycle trajectory with known bounds, so that the delay-free analysis of
the tracking law applies unchanged (Lemma~\ref{lem:wp}).
\item As without coupling, the errors converge for every bounded delay on directed, switching graphs, exponentially on
bounded sets with delay-independent constants (Theorems~\ref{thm:conv} and~\ref{thm:exp}).
\item On undirected graphs, the disagreement caused by persistent local errors below a design bound shrinks by a
delay-independent factor; for time-varying errors, delays degrade the attenuation (Proposition~\ref{prop:form},
Remark~\ref{rem:freq}).
\item The linearization of a direct coupling is stable for every delay if and only if the product of its gain and the
degree is below a threshold set by the tracking law (Proposition~\ref{prop:direct}). Simulations confirm this and
compare both laws with a law that has the gains but not the data of the direct coupling (Section~\ref{sec:sim}).
\end{enumerate}
Proofs are sketched; full proofs are given in \proofsin.

\emph{Notation:} $|\cdot|$ is the Euclidean norm, $\otimes$ the Kronecker product, $\mathbf 1$ the vector of ones,
$(x)_+=\max\{x,0\}$, $D^{+}$ the upper right Dini derivative, $\xi(\theta)=(\cos\theta,\sin\theta)^{\top}$,
$\xip(\theta)=(-\sin\theta,\cos\theta)^{\top}$ and $\sinc\theta=\sin\theta/\theta$ with $\sinc0=1$.

%%%%%%%%%%%%%%%%%%%%%%%%%%%%%%%%%%%%%%%%%%%%%%%%%%%%%%%%%%%%%%%%%%%%%%%%%%%%%%%%
\section{PROBLEM FORMULATION}\label{sec:prob}

AGV $i\in\{1,\dots,N\}$ has position $p_i\in\R^2$ and heading $\theta_i$ and obeys
\begin{equation}\label{eq:model}
\dot p_i=v_i\,\xi(\theta_i),\qquad \dot\theta_i=\omega_i ,
\end{equation}
with the forward speed $v_i$ and the turn rate $\omega_i$ as inputs. A planner assigns to each vehicle a reference
$(p_{ri},\theta_{ri})$ that obeys the same kinematics with speed $v_{ri}$ and turn rate $\omega_{ri}$. The desired
formation is given by the offsets $p_{ri}-p_{rj}$, which may vary with time. With the tracking error $e_i=p_{ri}-p_i$,
the formation error of vehicles $i$ and $j$ is $(p_i-p_j)-(p_{ri}-p_{rj})=e_j-e_i$.

AGV $j$ periodically broadcasts a vector $\sigma_j\in\R^2$, defined in \eqref{eq:local} below. At time $t$, AGV $i$
holds the newest sample it has received from $j$, denoted by $\sigma_j(t-\tau_{ij}(t))$; outdated packets are discarded.
The weight $a_{ij}(t)\ge0$ specifies how strongly AGV $i$ uses the data of AGV $j$. Let $d_i(t)=\sum_ja_{ij}(t)$.

\begin{assumption}\label{A1}
$v_{\min}\le v_{ri}(t)\le v_{\max}$ with $v_{\min}>0$, $|\dot v_{ri}(t)|\le a_{\max}$ and
$|\omega_{ri}(t)|\le\omega_{\max}$; $\dot v_{ri}$ and $\omega_{ri}$ are piecewise continuous and known to vehicle $i$.
\end{assumption}
\begin{assumption}\label{A2}
The weights $a_{ij}(t)$ are piecewise constant and right-continuous with finitely many switches in every bounded
interval, $a_{ij}(t)=0$ until the first packet from $j$ has arrived at $i$, and $d_i(t)\le d$.
\end{assumption}
\begin{assumption}\label{A3}
On every link, $\tau_{ij}(t)\in[0,t]$ is piecewise continuous and $t-\tau_{ij}(t)$ is nondecreasing: $\tau_{ij}$
grows with unit rate between packet arrivals and drops at arrivals, which are finitely many in bounded time.
\end{assumption}
\begin{assumption}\label{A4}
(a) $t-\tau_{ij}(t)\to\infty$ on every link with $a_{ij}(t)\not\to0$; or (b) there is a finite $h$, not necessarily
known, such that $\tau_{ij}(t)\le h$ whenever $a_{ij}(t)>0$.
\end{assumption}

Neither symmetry nor connectivity of the graph is required, and Assumption~\ref{A3} does not bound the delays. Since
$t-\tau_{ij}$ does not decrease, Assumption~\ref{A4}(b) implies~\ref{A4}(a); \ref{A4}(b) covers sampling, jitter,
reordering and bounded bursts of packet loss, and \ref{A4}(a) also admits unbounded delays. Formations that stop are
not covered; all vehicles share a map frame and the planner's time base. The objectives are
(i)~$e_i\to0$ and $\theta_{ri}-\theta_i\to0$; (ii)~$e_i-e_j\to0$; and (iii)~a quantifiable reduction of $e_i-e_j$ under
persistent perturbations compared with no coupling.

%%%%%%%%%%%%%%%%%%%%%%%%%%%%%%%%%%%%%%%%%%%%%%%%%%%%%%%%%%%%%%%%%%%%%%%%%%%%%%%%
\section{CONTROLLER DESIGN}\label{sec:ctrl}

\begin{figure}[t]
\centering
\begin{tikzpicture}[>={Stealth[length=1.5mm]}, font=\scriptsize,
  blk/.style={draw, rounded corners=1.2pt, align=center, inner sep=1.5pt, minimum height=8mm}]
\node[blk, dashed, text width=12.6mm] (net) {Network\\[1pt] $\sigma_j(t{-}\tau_{ij})$};
\node[blk, right=4.2mm of net, fill=black!8, text width=11.4mm] (filt) {Filter \eqref{eq:filter}\\[1pt] $z_i,w_i,u_i$};
\node[blk, right=4.2mm of filt, text width=15.2mm] (ref) {Shifted ref.\ \eqref{eq:shifted}\\[1pt] $\hat p_i,\vh_i,\wh_i$};
\node[blk, right=4.2mm of ref, text width=8.8mm] (law) {Law \eqref{eq:law}\\[1pt] $v_i,\omega_i$};
\node[blk, right=4.2mm of law, text width=10mm] (agv) {AGV \eqref{eq:model}\\[1pt] $p_i,\theta_i$};
\node[above=2.4mm of ref, align=center] (plan) {planner: $p_{ri},\theta_{ri},v_{ri},\dot v_{ri},\omega_{ri}$};
\draw[->] (net) -- (filt);
\draw[->] (filt) -- (ref);
\draw[->] (plan) -- (ref);
\draw[->] (ref) -- (law);
\draw[->] (law) -- (agv);
\draw[->] (agv.south) -- ++(0,-3mm) -| node[pos=0.25,below=-0.4mm]{pose} ($(law.south)+(2mm,0)$);
\draw[->] ($(law.south)+(-2mm,0)$) -- ++(0,-3mm) -| node[pos=0.25,below=-0.4mm]{$\eps_i=\hat p_i-p_i$} ($(filt.south)+(2mm,0)$);
\draw[->] ($(filt.south)+(-2mm,0)$) -- ++(0,-3mm) -| node[pos=0.25,below=-0.4mm]{$\sigma_i$} (net.south);
\end{tikzpicture}
\caption{Controller of AGV $i$. Delayed neighbor data enter only the reference-shift filter.}\label{fig:cascade}
\end{figure}
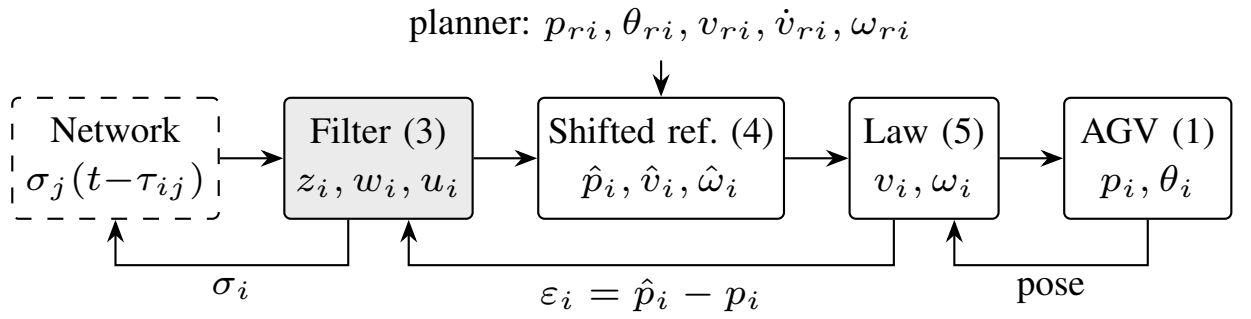

The reference-shift filter of AGV $i$ (Fig.~\ref{fig:cascade}) has two states $z_i,w_i\in\R^2$, the shift of the
vehicle's reference and the rate of the shift. From them, the vehicle forms the shifted reference position $\hat p_i$,
the local error $\eps_i$, and the vectors $\zeta_i$ and $\sigma_i$,
\begin{equation}\label{eq:local}
\begin{aligned}
&\hat p_i=p_{ri}-z_i, && \eps_i=\hat p_i-p_i,\\
&\zeta_i=z_i+w_i/\gamma, && \sigma_i=\zeta_i+\eps_i,
\end{aligned}
\end{equation}
and broadcasts $\sigma_i$. Note that $e_i=z_i+\eps_i$ and $\sigma_i=e_i+w_i/\gamma$. The filter reads
\begin{equation}\label{eq:filter}
\begin{aligned}
g_i&=-\lambda\zeta_i-\Lambda\sum_ja_{ij}\bigl[\zeta_i+\phi(\eps_i)-\sigma_j(t-\tau_{ij})\bigr],\\
u_i&=g_i\min\{1,\rho v_{ri}/|g_i|\},\\
\dot w_i&=\gamma(u_i-w_i),\qquad \dot z_i=w_i,
\end{aligned}
\end{equation}
with $\phi(\eps)=\eps\min\{1,\bar e/|\eps|\}$. Here $\Lambda\ge0$ is the coupling gain, $\lambda>0$ the restoring gain
that returns the shift to zero, $\gamma>0$ the filter bandwidth, $\rho\in(0,1)$ the largest commanded shift speed
$|u_i|$ relative to the reference speed, and $\bar e>0$ bounds the vehicle's own error inside the filter, so that a
vehicle that cannot follow its reference does not keep shifting it. By \eqref{eq:local} and \eqref{eq:filter},
$\dot\zeta_i=u_i$.

The velocity and the acceleration of the shifted reference are available from local quantities,
\begin{equation}\label{eq:shifted}
\begin{aligned}
\vvh_i&=v_{ri}\,\xi(\theta_{ri})-w_i,\\
\avh_i&=\dot v_{ri}\,\xi(\theta_{ri})+v_{ri}\,\omega_{ri}\,\xip(\theta_{ri})-\gamma(u_i-w_i).
\end{aligned}
\end{equation}
They define a unicycle trajectory with speed $\vh_i=|\vvh_i|$, heading $\thh_i=\arg\vvh_i$ (continuous lift,
$\thh_i(0)=\theta_{ri}(0)$) and turn rate $\wh_i=(\hat{\mathrm v}_{i,x}\hat{\mathrm a}_{i,y}-\hat{\mathrm
v}_{i,y}\hat{\mathrm a}_{i,x})/\vh_i^{2}$. The filter is of second order so that the received samples enter $u_i$, which
enters $\wh_i$ algebraically; no received signal has to be differentiated. The gain-scheduled law of
\cite{samson1993,deluca2001} tracks the shifted reference:
\begin{equation}\label{eq:law}
\begin{aligned}
&\mu_i=\xi(\theta_i)^{\top}\eps_i,\quad \nu_i=\xip(\theta_i)^{\top}\eps_i,\quad \vt_i=\thh_i-\theta_i,\\
&k_i=2\psi\sqrt{\wh_i^{2}+b\,\vh_i^{2}},\qquad v_i=\vh_i\cos\vt_i+k_i\mu_i,\\
&\omega_i=\wh_i+k_i\vt_i+b\,\vh_i\sinc(\vt_i)\,\nu_i ,
\end{aligned}
\end{equation}
with gains $\psi,b>0$. For $\Lambda=0$ the filter stays at rest, and \eqref{eq:law} is the single-vehicle law that
tracks the planner's reference (\emph{no coupling}). Neither the delays nor a bound on them is used.

\begin{assumption}\label{A5}
$\lambda,\psi,b,\bar e>0$, $\Lambda\ge0$, $0<\rho<1$ and $\gamma>\rho\,a_{\max}/((1-\rho)v_{\min})$.
\end{assumption}

%%%%%%%%%%%%%%%%%%%%%%%%%%%%%%%%%%%%%%%%%%%%%%%%%%%%%%%%%%%%%%%%%%%%%%%%%%%%%%%%
\section{STABILITY ANALYSIS}\label{sec:ana}

We use the vehicle Lyapunov function $W_i=\tfrac12|\eps_i|^{2}+\vt_i^{2}/(2b)$ and the constants
\begin{gather*}
\hat v_{\min}=(1-\rho)v_{\min}-\rho a_{\max}/\gamma,\qquad \bar\kappa=\bar A/\hat v_{\min}^{2},\\
\bar A=\sqrt{a_{\max}^{2}+v_{\max}^{2}\omega_{\max}^{2}}+\rho(2\gamma v_{\max}+a_{\max}),
\end{gather*}
where $\hat v_{\min}>0$ by Assumption~\ref{A5}. We say that the filters start at rest if $z_i(0)=w_i(0)=0$.

\subsection{Feasibility and Delay-Independent Convergence}

\begin{lemma}\label{lem:wp}
Let Assumptions \ref{A1}--\ref{A3} and \ref{A5} hold and let the filters start at rest. For all initial poses and all
delays, the closed loop \eqref{eq:model}--\eqref{eq:law} has a unique solution on $[0,\infty)$, and for all $t\ge0$,
$\vh_i\ge\hat v_{\min}$, $|\avh_i|\le\bar A$, $|\wh_i|\le\bar\kappa\,\vh_i$, $|\thh_i-\theta_{ri}|<\pi/2$ and
$|\sin(\thh_i-\theta_{ri})|\le|w_i|/v_{ri}$.
\end{lemma}

\begin{proof}[Proof (sketch)]
As $|u_i|\le\rho v_{ri}$ and $|\dot v_{ri}|\le a_{\max}$, $y_i=|w_i|-\rho v_{ri}-\rho a_{\max}/\gamma$ obeys
$D^{+}y_i\le-\gamma y_i$, so $|w_i|\le\rho(v_{ri}+a_{\max}/\gamma)<v_{ri}$, which gives the bounds. Between arrivals,
the closed loop is locally Lipschitz where $|w_i|<v_{ri}$, and $W_i$ does not increase (Theorem~\ref{thm:conv}(i) uses
only \eqref{eq:law}), so there is no finite escape time.
\end{proof}

The shifted reference is thus kinematically feasible, which addresses the issue raised in \cite{maghenem2021lcss};
its turn rate, however, jumps at packet arrivals.

\begin{theorem}\label{thm:conv}
Let Assumptions \ref{A1}--\ref{A3} and \ref{A5} hold, let the filters start at rest, and let
$E_0=\max_j\sqrt{2W_j(0)}$. Then, for all initial poses:
\begin{enumerate}
\item[(i)] $\dot W_i=-k_i\mu_i^{2}-(k_i/b)\vt_i^{2}\le0$ for almost all $t$, for any delays and any graph;
\item[(ii)] for all $t\ge0$ and arbitrary delays, $|\eps_i(t)|\le E_0$, $|\zeta_i(t)|\le(2\Lambda d/\lambda)E_0$ and
$|e_i(t)|\le CE_0$ with $C=1+2\Lambda d/\lambda+(4\Lambda d/\gamma)(1+\Lambda d/\lambda)$;
\item[(iii)] under Assumption~\ref{A4}(a), $e_i(t)\to0$ and $\theta_{ri}(t)-\theta_i(t)\to0$ for all $i$; hence
$e_i(t)-e_j(t)\to0$ for all $i,j$.
\end{enumerate}
\end{theorem}

\begin{proof}
(i) The shifted reference is a unicycle trajectory with velocities $(\vh_i,\wh_i)$. Inserting \eqref{eq:law} gives,
with the index $i$ dropped,
\begin{equation}\label{eq:loop}
\dot\mu=\omega\nu-k\mu,\ \ \dot\nu=-\omega\mu+\vh\sin\vt,\ \ \dot\vt=-k\vt-b\vh\sinc(\vt)\nu,
\end{equation}
and the cross terms cancel in $\dot W_i$, as for a single vehicle \cite{samson1993}.

(ii) By (i), $|\eps_i|\le E_0$. With $\sigma_j=\zeta_j+\eps_j$, \eqref{eq:filter} becomes
\begin{equation}\label{eq:net}
\begin{aligned}
\dot\zeta_i&=\chi_i\Bigl[-(\lambda+\Lambda d_i)\zeta_i+\Lambda\sum_ja_{ij}\zeta_j(t-\tau_{ij})+\varpi_i\Bigr],\\
\varpi_i&=-\Lambda d_i\,\phi(\eps_i)+\Lambda\sum_ja_{ij}\,\eps_j(t-\tau_{ij}),
\end{aligned}
\end{equation}
where $\chi_i=\min\{1,\rho v_{ri}/|g_i|\}\in(0,1]$ and $|\varpi_i|\le\bar\varpi:=2\Lambda dE_0$. As
$t-\tau_{ij}\in[0,t]$, the delayed terms are bounded by $\bar M(t)=\max_{s\le t}M(s)$, $M(t)=\max_i|\zeta_i(t)|$, so
$D^{+}|\zeta_i|\le\chi_i[-\lambda|\zeta_i|+\Lambda d(\bar M-|\zeta_i|)+\bar\varpi]$ wherever $\zeta_i\ne0$. Where $M$
would first exceed some $\upsilon>\bar\varpi/\lambda$, this bracket is negative for the largest $|\zeta_i|$, so
$M\le\bar\varpi/\lambda$ whatever the delays: the self-damping $\lambda+\Lambda d_i$ exceeds the delayed weight
$\Lambda d_i$, as in \cite{munz2011,peng2023}. Then $|g_i|\le\bar g:=4\Lambda d(1+\Lambda d/\lambda)E_0$,
$|w_i|\le\bar g$, $|e_i|\le|\eps_i|+|\zeta_i|+|w_i|/\gamma\le CE_0$, and $\chi_i\ge\chi_0:=\min\{1,\rho v_{\min}/\bar g\}$.

(iii) As $k_i\ge2\psi\sqrt b\,\hat v_{\min}$, (i) gives $\mu_i,\vt_i\in L_2$ with bounded derivatives, so
$\mu_i,\vt_i\to0$, and a Barbalat-type argument on $\dot\vt_i$ gives $\nu_i\to0$. Hence $\eps_i\to0$ and, under
Assumption~\ref{A4}(a), $\varpi_i\to0$. The argument of (ii), applied to limits superior, gives
$\limsup M\le\limsup_t\max_i|\varpi_i|/\lambda=0$, so $\zeta_i\to0$, then $w_i,z_i,e_i\to0$, and
$\theta_{ri}-\theta_i\to0$ by Lemma~\ref{lem:wp}.
\end{proof}

Theorem~\ref{thm:conv} also holds for $\Lambda=0$; it shows only that delayed coupling cannot destroy tracking.

\subsection{Exponential Convergence}

A cross term $\nu\vt$ supplies the decrease that $\dot W_i$ lacks in $\nu_i$, as in the standard construction of strict
Lyapunov functions \cite{malisoff2009}; for unicycle tracking see also \cite{maghenem2018,xu2024}. For a length scale
$\ell>0$ and a curvature bound $\kappa>0$, let
\[
U_\ell=\bigl(1+\alpha_\ell\sqrt b/4\bigr)W+\alpha_\ell\,\nu\vt/\bigl(1+W/\ell^{2}\bigr),
\]
where $\alpha_\ell=\ell\,\alpha(\psi,b\ell^2,\kappa\ell)$, $c_\alpha=\tfrac13+\tfrac1{2\sqrt2}$ and
\begin{gather*}
\alpha(\psi,b,\kappa)=\min\bigl\{2\psi/(1+\sqrt{b/2}),\ 1/B\bigr\},\\
B=2\psi\sqrt{\kappa^{2}+b}+\sqrt2\,b+\tfrac{b}{\psi}\bigl(c_\alpha\sqrt b+\tfrac12+b^{-1/2}\bigr).
\end{gather*}

\begin{lemma}\label{lem:strict}
Along \eqref{eq:loop} with $|\wh|\le\kappa\vh$, $c_1W\le U_\ell\le c_2W$ with $c_1=1-\tfrac34\alpha_\ell\sqrt b\ge0.73$
and $c_2=1+\tfrac54\alpha_\ell\sqrt b\le1.45$, and $\dot U_\ell\le-\alpha_\ell\,b\,\vh\,W/(1+W/\ell^{2})$ for almost
all $t$.
\end{lemma}

By Lemma~\ref{lem:wp}, $\kappa=\bar\kappa$ is admissible in the closed loop; $\alpha_\ell$, $c_1$ and $c_2$ refer to
$\bar\kappa$ unless stated otherwise.

\begin{lemma}\label{lem:halanay}
Consider \eqref{eq:net} with measurable $\chi_i(t)\in[\chi_0,1]$, $\chi_0>0$, measurable, locally bounded $\varpi_i$,
$d_i\le d$ and $\tau_{ij}(t)\le\min\{t,h\}$ on active links, and let $M_h(t)=\max\{M(s):\max\{0,t-h\}\le s\le t\}$ and
$P(t)=\max_i|\varpi_i(t)|$. For $\beta\in(0,1)$, let $\eta>0$ solve $\eta=\chi_0[(1-\beta)\lambda-\Lambda
d(\ee^{\eta h}-1)]$. Then, for all $t\ge t_0\ge0$,
\[
M(t)\le\max\Bigl\{M_h(t_0)\,\ee^{-\eta(t-t_0)},\ \frac{1}{\beta\lambda}\sup_{s\in[t_0,t]}P(s)\Bigr\}.
\]
\end{lemma}

\begin{theorem}\label{thm:exp}
Let Assumptions \ref{A1}--\ref{A3}, \ref{A4}(b) and \ref{A5} hold, let the filters start at rest, and fix $\ell>0$.
Then $|\eps_i(t)|\le C_1E_0\ee^{-\eta_1t}$, $|e_i(t)|\le KE_0\ee^{-\eta^*t}$ and
$|\theta_{ri}(t)-\theta_i(t)|\le K'E_0\ee^{-\eta^*t}$ for all $i$ and $t\ge0$, with $\eta_1=\alpha_\ell b\hat
v_{\min}/(2c_2)$, $C_1=\sqrt{c_2/c_1}\,\ee^{E_0^{2}/(4\ell^{2})}$, $\eta^*=\tfrac12\min\{\eta_1,\eta_2,\gamma\}$, where
$\eta_2$ is the rate of Lemma~\ref{lem:halanay} for $\beta=\tfrac12$ and $\chi_0$ of Theorem~\ref{thm:conv}, and
constants $K,K'$ that do not depend on $h$, given in \proofsin.
\end{theorem}

In Lemma~\ref{lem:strict}, the two entries of $\alpha$ keep the other terms below the decrease supplied by the cross
term; Lemma~\ref{lem:halanay} is a max-norm Halanay inequality with an input \cite{halanay1966,wen2008} whose rate
equation, for $\chi_0=1$, $\beta\to0$, is that of positive delayed systems \cite{feyz2014}. Proofs: \proofsin.

\begin{remark}\label{rem:uas}
The delays enter only the rate $\eta^*$, through $h$, but the rate and $C_1$, $K$, $K'$ depend on $E_0$. With the a priori bound $\bar\kappa$, $\eta_1$ is tiny
(Section~\ref{sec:sim}), but only transiently: by Theorem~\ref{thm:conv}(iii), $g_i,w_i\to0$, so
$\wh_i-\omega_{ri}\to0$ and $\vh_i-v_{ri}\to0$, and from a finite time on, $|\eps_i|$ decays at every rate below
$\eta_1^\infty=\alpha_\ell bv_{\min}/(2c_2)$, with $\alpha_\ell$, $c_2$ for the curvature bound
$\kappa_r=\sup_{i,t}|\omega_{ri}|/v_{ri}$ of the planner's references. By Lemma~\ref{lem:strict}, velocity
perturbations also obey a local input-to-state estimate with delay-independent gains; see \corref.
\end{remark}

\subsection{Benefit of the Coupling}

\begin{proposition}\label{prop:form}
Let Assumptions \ref{A1}--\ref{A3}, \ref{A4}(a) and \ref{A5} hold, the filters start at rest, and the weights be
constant after some finite time, with Laplacian $L=D-A$, $A=[a_{ij}]$, $D=\operatorname{diag}(d_i)$. Under measurable,
locally bounded velocity perturbations in \eqref{eq:model}, suppose that a solution exists on $[0,\infty)$ (as for small
perturbations, \corref) along which $\eps_i(t)\to\eps_i^\infty$.
Then $e(t)\to e^\infty$, where $\phi$ acts on each block and
\begin{equation}\label{eq:form}
\begin{aligned}
e^\infty&=\bigl[(\lambda I+\Lambda L)^{-1}\otimes I_2\bigr]\,q^\infty,\\
q^\infty&=\lambda\eps^\infty+\Lambda(D\otimes I_2)\bigl(\eps^\infty-\phi(\eps^\infty)\bigr).
\end{aligned}
\end{equation}
If the graph is undirected and connected, with algebraic connectivity $\lambda_2(L)$, and
$\Pi=I-\tfrac1N\mathbf1\mathbf1^{\top}$, then
\[
\bigl|(\Pi\otimes I_2)e^\infty\bigr|\le\frac{\lambda\bigl|(\Pi\otimes I_2)\eps^\infty\bigr|+\Lambda d\,\bigl|\bigl((|\eps_i^\infty|-\bar e)_+\bigr)_i\bigr|}{\lambda+\Lambda\lambda_2(L)},
\]
and if $|\eps_i^\infty|\le\bar e$ for all $i$, then $\sum_ie_i^\infty=\sum_i\eps_i^\infty$ and
$|(\Pi\otimes I_2)e^\infty|\le\lambda|(\Pi\otimes I_2)\eps^\infty|/(\lambda+\Lambda\lambda_2(L))$. None of these
limits depends on the delays.
\end{proposition}

\begin{proof}
$\lambda I+\Lambda L$ is strictly diagonally dominant and thus nonsingular. As $\eps$ converges, it is bounded, and the
argument of Theorem~\ref{thm:conv}(ii) bounds $M$ and keeps $\chi_i$ away from zero. Once the weights are constant,
$\zeta-\zeta^*$, with $\zeta^*$ the equilibrium of \eqref{eq:net} for the limit input, obeys \eqref{eq:net} with an
input that tends to zero, so $\zeta\to\zeta^*$ by the limit-superior bound of Theorem~\ref{thm:conv}(iii). Then $w\to0$ and $e^\infty=\eps^\infty+\zeta^*$, which is \eqref{eq:form}. For an
undirected connected graph, $\Pi$ commutes with $(\lambda I+\Lambda L)^{-1}$, whose norm on the range of $\Pi$ is
$1/(\lambda+\Lambda\lambda_2(L))$; with $|\eps_i-\phi(\eps_i)|=(|\eps_i|-\bar e)_+$, this gives the bound.
\end{proof}

For $\Lambda=0$, $e^\infty=\eps^\infty$; the comparison presumes that the persistent local errors do not depend on the
coupling, as for constant speed perturbations on a straight reference. Below $\bar e$, the coupling
shrinks the disagreement $|(\Pi\otimes I_2)e|=(\tfrac1N\sum_{i<j}|e_i-e_j|^2)^{1/2}$, though not necessarily every
pairwise error, to at most $\lambda/(\lambda+\Lambda\lambda_2(L))$ times its value without coupling: if one vehicle
permanently loses speed, the others fall back. Errors above $\bar e$ reach the neighbors in full but the vehicle's own filter only through
$\phi(\eps_i)$, so the coupling can amplify the disagreement (by 25\% for $\lambda=\Lambda=1$\,s$^{-1}$,
$\bar e=5$\,cm and a 0.2\,m error at a vehicle off the edge joining two complete graphs of five); $\bar e$ should exceed
the persistent errors to be expected.

\begin{remark}\label{rem:freq}
Below $\bar e$ and with $\chi_i=1$, \eqref{eq:net} is linear; for constant weights and a uniform delay $\tau$, Laplace
transforms give $e=(H(s)\otimes I_2)\,\eps$ with $A_\tau=\ee^{-s\tau}A$ and
\begin{equation}\label{eq:H}
H(s)=I-\Lambda\bigl[(1+s/\gamma)(sI+\lambda I+\Lambda D-\Lambda A_\tau)\bigr]^{-1}(D-A_\tau),
\end{equation}
which is stable for every $\tau$. $H(0)=\lambda(\lambda I+\Lambda L)^{-1}$ gives Proposition~\ref{prop:form}, but
$|H(j\omega)|$ depends on $\tau$ for $\omega>0$: on the complete graph of three, where $H$ acts on disagreements as a
scalar $h_3$, $\lambda=\Lambda=1$\,s$^{-1}$ and $\tau=0$--3\,s give $|h_3(j\omega)|=0.26$--0.27 at 0.175\,rad/s and
0.37--0.56 at 0.68\,rad/s instead of 0.25, and up to 1.21 above 2--3.5\,rad/s. Local errors that rotate with the
reference, as on a circle, are thus attenuated less.
\end{remark}

\subsection{A Direct Coupling for Comparison}

In an unsaturated direct coupling \cite{kostic2010,sadowska2011,sadowska2012}, the vehicles broadcast $e_i$, and
$s_i=e_i+c\sum_ja_{ij}[e_i-e_j(t-\tau_{ij})]$, $c>0$, replaces $\eps_i$ in \eqref{eq:law} with the planner's reference;
this multiplies the gains acting on $e_i$ by $1+cd_i$, and stability is known up to a delay bound that shrinks as $c$
grows \cite{alvarez2011}.

\begin{proposition}\label{prop:direct}
For a straight reference with constant speed $v$, an undirected $d$-regular graph and a uniform delay $\tau$, the
linearized direct coupling has longitudinal errors that are stable for every $\tau\ge0$ and lateral errors
$e_\perp\in\R^N$ that obey, with $k=2\psi\sqrt b\,v$,
\begin{equation}\label{eq:lin}
\ddot e_\perp+k\dot e_\perp+bv^2\bigl[(1+cd)e_\perp-cA\,e_\perp(t-\tau)\bigr]=0 .
\end{equation}
These are stable for every $\tau\ge0$ if and only if $cd<2\psi(1+\psi)$; otherwise the mode
$e_\perp\in\mathrm{span}\{\mathbf1\}$ becomes unstable at a delay proportional to $1/v$.
\end{proposition}

\begin{proof}[Proof (sketch)]
An eigenvalue $a\in[-d,d]$ of $A$ gives the mode $s^2+ks+bv^2(1+cd-ca\,\ee^{-s\tau})$; by the frequency-sweeping test
\cite{gu2003}, it is stable for all $\tau$ if and only if $|(j\omega)^2+jk\omega+bv^2(1+cd)|>bv^2c|a|$ for all $\omega$,
and the minimum of the left side exceeds $bv^2cd$ if and only if $cd<2\psi(1+\psi)$.
\end{proof}

%%%%%%%%%%%%%%%%%%%%%%%%%%%%%%%%%%%%%%%%%%%%%%%%%%%%%%%%%%%%%%%%%%%%%%%%%%%%%%%%
\section{SIMULATION RESULTS}\label{sec:sim}

\begin{figure}[t]
\centering
\includegraphics[width=\columnwidth]{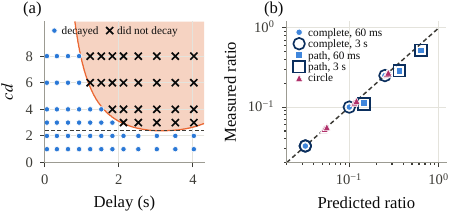}
\caption{Kinematic model. (a) Direct coupling, straight reference at 0.6\,m/s, complete graph of three, constant
delay: region where \eqref{eq:lin} is unstable (shaded; dashed: $cd=2\psi(1+\psi)$), runs whose tracking errors
decayed (dots) or not (crosses).
(b) Steady disagreement with over without coupling for a persistent speed loss, against the factor of
Proposition~\ref{prop:form} (straight; complete, $N=3$; path, $N=4$) and $|h_3(j\omega_r)|$ (circle).}\label{fig:kin}
\end{figure}

\begin{figure*}[t]
\centering
\includegraphics[width=\textwidth]{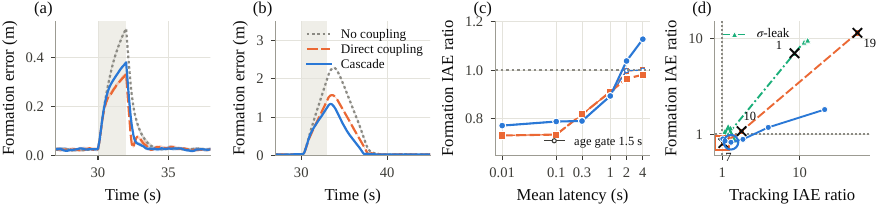}
\caption{Detailed AGV model, warehouse route. (a) Traction loss of AGV~2 at 30\,s (40\% of the commanded velocities
for 2\,s) and (b) safety stop of AGV~2 for 3\,s (shaded): formation error, median and interquartile range over 20 seeds.
(c) Formation IAE after the traction loss relative to no coupling against the nominal mean latency (ratio of medians,
10 seeds); thin lines: age gate of 1.5\,s. (d) Gain selection on separate tuning seeds: whole-run IAE relative to no
coupling, one point per gain; circles: selected; crosses: gains with failed runs (number out of 20).}\label{fig:hifi}
\end{figure*}

Besides no coupling and the direct coupling, the cascade is compared with a \emph{high-gain law}, in which
$s_i=(1+cd_i)e_i$ replaces $\eps_i$ in \eqref{eq:law}, with the gains of the direct coupling but no neighbor data, and
with a \emph{$\sigma$-leak filter}, which transfers the filter of \cite{peng2023} to unicycles: its restoring term acts
on $\sigma_i$, and $\eps_i$ enters without $\phi$.

\subsection{Kinematic Model}

An event-driven simulation of exactly the model of Sections~\ref{sec:prob} and~\ref{sec:ctrl} checks every bound. Three AGVs keep a triangle along a circle, a figure-8
and a warehouse route at 0.15--1.10\,m/s ($\psi=0.7$, $b=3$\,m$^{-2}$, $\lambda=\Lambda=1$\,s$^{-1}$,
$\gamma=5$\,s$^{-1}$, $\rho=0.6$, $\bar e=0.05$\,m), over links with 20\,Hz broadcasts, random latency and bursty
loss (delays up to 4.16\,s). No bound was violated in 280 runs on complete, directed-chain and empty graphs with
initial errors up to 3\,m and $175^\circ$. The explicit rates are conservative: $\bar\kappa=3183$--6521\,m$^{-1}$ gives
$\eta_1\approx10^{-5}$\,s$^{-1}$, and $\kappa_r=0.28$--2.1\,m$^{-1}$ gives $\eta_1^\infty=0.012$--0.016\,s$^{-1}$
($\ell=1$\,m), whereas $\max_i|\eps_i|$ decays at 0.43--1.11\,s$^{-1}$.

For Proposition~\ref{prop:form}, AGV~1 permanently loses 0.03\,m/s of speed. On a straight reference, the errors after
150\,s agree with \eqref{eq:form} to $10^{-9}$\,mm at 60\,ms mean latency and to 0.21\,mm at 3\,s,
where they settle more slowly, and the disagreement falls by the predicted factors on the complete graph and by more on
a path (Fig.~\ref{fig:kin}(b)). On a circle with constant delays, it falls by $|h_3(j\omega_r)|$ of
Remark~\ref{rem:freq} to within $10^{-4}$, less than Proposition~\ref{prop:form} predicts.

For Proposition~\ref{prop:direct}, Fig.~\ref{fig:kin}(a) varies the gain of the direct coupling up to $cd=8$: in all
288 runs, its tracking errors decayed where \eqref{eq:lin} is stable and did not where it is unstable, whereas those of
the cascade decayed in all 144 runs, with $\Lambda=1$, 4 and 8\,s$^{-1}$ and delays up to 4\,s.

\subsection{Detailed AGV Model}

The simulator of a generic differential-drive AGV (150\,kg plus 0--100\,kg payload) adds 50\,Hz control on a
drifting clock; LiDAR-type fixes (25\,Hz, 40\,ms latency, 10\,mm noise and bias) fused with odometry; wheel-speed
setpoints with 1\% calibration errors, limited to 1.5\,m/s and 1\,m/s$^2$, delayed by 10\,ms and tracked by PI loops
with torques up to 20\,N\,m; rigid-body dynamics with slip; and links with 25\,Hz broadcasts, $30\pm15$\,ms latency
with spikes and 4\% bursty loss; $\vt_i$ is wrapped to $(-\pi,\pi]$. The coupling gains were tuned on separate seeds to the lowest formation IAE without failures:
$\Lambda=1$\,s$^{-1}$ for both filters and $c=0.5$, which the high-gain law shares (Fig.~\ref{fig:hifi}(d)). Unless
stated otherwise, three AGVs on a complete graph follow the warehouse route; a seed fixes the initial errors, vehicle
parameters, network, noise and slip. The formation error is $\max_{i<j}|e_i-e_j|$ and
the tracking error $\sum_i|e_i|/N$, with true positions; RMS values are taken over $t\ge15$\,s before any event, IAE is
the integral over 15\,s from the onset of an event, which affects AGV~2 at 30\,s (Table~\ref{tab:hifi}), and a run fails
when a tracking-error coordinate exceeds 3\,m.

\begin{table}[t]
\caption{Detailed AGV model: means over 30 (nominal) and 20 seeds (events); random draws: median and 95th percentile of whole-run IAE ratios to no coupling; tuning gains: $c=0.1$--4, $\Lambda=0.25$--8\,s$^{-1}$, 20 runs each. $^\dagger$: differs from the cascade, $p<0.01$ (paired Wilcoxon, Holm over four comparisons; random draws: log ratios)}\label{tab:hifi}
\centering\footnotesize\renewcommand{\arraystretch}{0.94}
\setlength{\tabcolsep}{2.0pt}
\begin{tabular}{@{}lccccc@{}}
\toprule
 & None & High g. & $\sigma$-leak & Direct & Cascade\\
\midrule
Nominal: form.\ RMS (mm) & 28.2 & 27.8 & 49.1$^\dagger$ & 28.5$^\dagger$ & 27.7\\
\quad RMS of $\dot v_i$ (m/s$^2$) & 0.70$^\dagger$ & 1.41$^\dagger$ & 0.68$^\dagger$ & 1.48$^\dagger$ & 0.65\\
\quad heading RMS (deg) & 0.55$^\dagger$ & 0.87$^\dagger$ & 2.61$^\dagger$ & 0.98$^\dagger$ & 1.07\\
Traction loss: peak (mm) & 517$^\dagger$ & 447$^\dagger$ & 649$^\dagger$ & 331$^\dagger$ & 380\\
\quad form.\ IAE (m$\cdot$s) & 1.24$^\dagger$ & 1.07$^\dagger$ & 3.91$^\dagger$ & 0.91$^\dagger$ & 0.98\\
\quad track.\ IAE (m$\cdot$s) & 0.52$^\dagger$ & 0.45$^\dagger$ & 2.46$^\dagger$ & 0.60$^\dagger$ & 0.93\\
Safety stop: peak (mm) & 2287$^\dagger$ & 2281$^\dagger$ & 1407$^\dagger$ & 1574$^\dagger$ & 1341\\
\quad form.\ IAE (m$\cdot$s) & 10.06$^\dagger$ & 10.03$^\dagger$ & 9.16$^\dagger$ & 6.83$^\dagger$ & 5.35\\
\quad track.\ IAE (m$\cdot$s) & 3.47$^\dagger$ & 3.45$^\dagger$ & 9.91$^\dagger$ & 5.59$^\dagger$ & 6.78\\
Random draws: form. & 1$^\dagger$ & 0.93 & 2.34$^\dagger$ & 0.88 & 0.89\\
\quad form., 95th percentile & 1 & 1.10 & 16.85 & 2.33 & 1.24\\
\quad track. & 1$^\dagger$ & 0.92$^\dagger$ & 2.36$^\dagger$ & 1.00$^\dagger$ & 1.24\\
Failed: tuning gains & 0/20 & 16/120 & 1/120 & 36/120 & 0/120\\
\quad $N=8$, 12, unit weights & 0/12 & 11/12 & -- & 12/12 & 0/12\\
\quad $N=8$, 12, normalized & -- & 0/12 & -- & 0/12 & 0/12\\
\quad random draws & 0/150 & 0/150 & 19/150 & 3/150 & 0/150\\
\bottomrule
\end{tabular}
\end{table}

\emph{Nominal operation:} All laws except the $\sigma$-leak filter give a formation RMS of 27--29\,mm, set mainly by the
localization bias (without it, 13 instead of 26\,mm). The direct coupling and the high-gain law double the RMS of $\dot v_i$; the cascade does not, but
it doubles the heading RMS.

\emph{Disturbances:} After a traction loss, the cascade lowers the peak and the IAE of the formation error by 27\% and
21\%, the direct coupling by 36\% and 26\% (Fig.~\ref{fig:hifi}(a)), and the high-gain law, without any data, by 13\% each:
about half of the benefit of the direct coupling stems from its higher gain, and on every seed the data lower the
formation IAE by a further 15\% (direct) and 9\% (cascade). The tracking IAE rises by 80\% (cascade) and 15\% (direct). With intermittent links (on and off for 5\,s and 2\,s on
average), the formation IAE falls by only 13\% and 20\%, for the cascade no more than with the high-gain law. After a
3\,s safety stop, the cascade reduces it by 47\%, the direct coupling by 32\% and the high-gain law not at all
(Fig.~\ref{fig:hifi}(b)): as in coordinated path following \cite{ghabcheloo2009,li2012}, the others wait. The $\sigma$-leak filter winds up whenever a vehicle cannot follow its
reference, which $\phi$ and restoring $\zeta_i$ prevent; it did best only on the kinematic model with the delays of
\cite{peng2023}.

\emph{Delays:} Up to a mean latency of 0.3\,s, both laws reduce the formation IAE after the traction loss by 18--27\%,
at 1\,s by 9--11\%; at 2 and 4\,s, the ratio is 0.96 and 0.98 for the direct coupling and 1.04 and 1.13 for the
cascade, which was then worse on every seed (Fig.~\ref{fig:hifi}(c)). An age gate that discards samples
older than 1.5\,s, which Assumption~\ref{A2} admits and which makes $h$ in Assumption~\ref{A4}(b) a design choice, left
the ratios up to 1\,s unchanged and returned them to 1.00 at 2 and 4\,s (thin lines). No run failed for latencies up
to 4\,s, 90\% packet loss or 1\,Hz broadcasts; with $cd=1<2\psi(1+\psi)$, the tuned direct coupling is
delay-independently stable in its linearization (Proposition~\ref{prop:direct}).

\emph{Gains, team size and graphs:} The direct coupling failed for $c\ge1$ on the tuning seeds and with unit weights on
complete graphs of 8 and 12 vehicles, the high-gain law for $c\ge2$ (16 of 40 runs) and on these graphs (11 of 12):
the failures stem mainly from the higher gain, which saturates the drives (without drive limits, the direct coupling
did not fail at $c\le2$). The cascade never failed but degraded for $\Lambda\ge4$\,s$^{-1}$ and on these graphs. On
sparse graphs (rings of 5--12 vehicles; stars, directed rings and chains of 3--12), the direct coupling lowered the
formation IAE after the traction loss by 6--13\% and the cascade by at most 8\% (on directed chains, it raised it by
1--5\%). Over 150 random draws of all parameters, the median formation-IAE ratio is 0.88 (direct), 0.89 (cascade) and
0.93 (high gain); the cascade has a smaller tail than the direct coupling but a 24\% higher median tracking IAE.

\emph{When the coupling hurts:} Both laws pass measurement noise between vehicles: with 2\,cm of localization noise,
a 300\,kg payload, a 100\,ms actuation delay or 10\,Hz control, their median steady formation error is 20--60\% higher
than without coupling, and with 4\,cm of noise ten times as high for the direct coupling (cascade: $+49\%$).

%%%%%%%%%%%%%%%%%%%%%%%%%%%%%%%%%%%%%%%%%%%%%%%%%%%%%%%%%%%%%%%%%%%%%%%%%%%%%%%%
\section{CONCLUSIONS}

Keeping the delays out of the vehicle loop yields a formation-tracking controller for unicycle-type AGVs whose coupling
preserves tracking for any bounded delay and directed, switching graph and provably reduces the disagreement caused by
persistent local errors on undirected graphs, whereas a direct coupling is delay-independent only below a gain
threshold. In simulation, the cascade never failed and reduced the formation error after traction losses and safety
stops at a cost in tracking accuracy; the tuned direct coupling did better after traction losses, partly through its
higher gain. Long latencies call for an age gate, and on sparse graphs and larger teams the benefit was small; experiments
with physical AGVs are future work.

\ifarxiv
%%%%%%%%%%%%%%%%%%%%%%%%%%%%%%%%%%%%%%%%%%%%%%%%%%%%%%%%%%%%%%%%%%%%%%%%%%%%%%%%
% Appendix of the arXiv version: full proofs. Included by root.tex only when \ARXIVVERSION is defined.
\section*{APPENDIX}

\subsection{Proof of Lemma~\ref{lem:wp}}

By construction, $|u_i|\le\rho v_{ri}$. Let $y_i=|w_i|-\rho v_{ri}-\rho a_{\max}/\gamma$. From $\dot w_i=\gamma(u_i-w_i)$
and $|\dot v_{ri}|\le a_{\max}$ we obtain $D^{+}y_i\le\gamma(\rho v_{ri}-|w_i|)+\rho a_{\max}=-\gamma y_i$, so that
$y_i\le0$ is preserved; filters at rest satisfy $y_i(0)\le0$. Hence, for all $t$,
$\vh_i\ge v_{ri}-|w_i|\ge(1-\rho)v_{\min}-\rho a_{\max}/\gamma=\hat v_{\min}$, which is positive by
Assumption~\ref{A5}, and $|\avh_i|\le\sqrt{a_{\max}^2+v_{\max}^2\omega_{\max}^2}+\gamma(|u_i|+|w_i|)\le\bar A$. Since
$|\wh_i|\le|\avh_i|/\vh_i$, also $|\wh_i|/\vh_i\le\bar A/\hat v_{\min}^2=\bar\kappa$. Because $|w_i|<v_{ri}$, the vector
$\vvh_i=v_{ri}\xi(\theta_{ri})-w_i$ has a positive component along $\xi(\theta_{ri})$, so the continuous lift of
$\thh_i-\theta_{ri}$, which starts at $0$, stays in $(-\pi/2,\pi/2)$; the law of sines in the triangle with the sides
$v_{ri}\xi(\theta_{ri})$, $w_i$ and $\vvh_i$ gives $|\sin(\thh_i-\theta_{ri})|\le|w_i|/v_{ri}$.

Existence and uniqueness: With newest-sample hold, the received samples are constant between arrivals, and there are
finitely many arrivals, weight switches and discontinuities of $\dot v_{ri}$ and $\omega_{ri}$ in bounded time. On the
set $\{|w_i|<v_{ri}\}$, the lift is the smooth function
$\thh_i=\theta_{ri}+\arctan\bigl(\xip(\theta_{ri})^{\top}\vvh_i/\xi(\theta_{ri})^{\top}\vvh_i\bigr)$ of $\theta_{ri}$
and $w_i$, whose denominator is positive. On each interval between these events, the closed loop is therefore an
ordinary differential equation in $(p_i,\theta_i,z_i,w_i)$ whose right-hand side is piecewise continuous in $t$ and
locally Lipschitz in the state on $\{|w_i|<v_{ri}\}$ (the radial saturations, $\phi$, $\sinc$ and the square root in
$k_i$ are Lipschitz there, since $\vh_i>0$). On the maximal interval of existence, the bound on $w_i$ holds, because it
uses only $|u_i|\le\rho v_{ri}$, so the solution stays in this set; the identity of Theorem~\ref{thm:conv}(i), which uses
only the structure of \eqref{eq:law}, shows that $W_i$ does not increase and hence bounds $\eps_i$ and $\vt_i$; and
$|\dot z_i|=|w_i|$ is bounded. Hence there is no finite escape time, and the solution exists on $[0,\infty)$.
\hfill$\blacksquare$

\subsection{Proof of Theorem~\ref{thm:conv}(ii) and (iii)}

\emph{(ii):} Let $M(t)=\max_i|\zeta_i(t)|$ and $\bar M(t)=\max_{s\le t}M(s)$. As $t-\tau_{ij}\in[0,t]$, the delayed terms
of \eqref{eq:net} are bounded by $\bar M(t)$; with $\bar M\ge|\zeta_i|$ and $d_i\le d$, wherever $\zeta_i\ne0$,
\[
D^{+}|\zeta_i|\le\chi_i\bigl[-\lambda|\zeta_i|+\Lambda d\bigl(\bar M(t)-|\zeta_i|\bigr)+\bar\varpi\bigr],
\]
whose bracket is continuous in $t$. Suppose that $M$ exceeds some $\upsilon>\bar\varpi/\lambda$ right after $t_1$, with
$M\le\upsilon$ on $[0,t_1]$. At $t_1$ the bracket equals $-\lambda\upsilon+\bar\varpi<0$ for every $i$ with
$|\zeta_i(t_1)|=\upsilon$, and it stays negative shortly after $t_1$, so $M\le\upsilon$ there, a contradiction. Hence
$M\le\bar\varpi/\lambda$. Then $|g_i|\le(\lambda+\Lambda d)M+\Lambda d\bar M+\bar\varpi\le\bar g$, $|w_i|\le\bar g$ by
$\dot w_i=\gamma(u_i-w_i)$ and $w_i(0)=0$, and the remaining bounds follow from $e_i=\zeta_i-w_i/\gamma+\eps_i$.

\emph{(iii):} We use the following Barbalat-type lemma.

\begin{lemma}\label{lem:barb}
Let $f$ be absolutely continuous on $[0,\infty)$ with a finite limit, $\dot f=q+r$ almost everywhere, $q$ uniformly
continuous and $r(t)\to0$. Then $q(t)\to0$.
\end{lemma}

\begin{proof}
Otherwise, there are $m_0>0$ and $t_n\to\infty$ with $|q(t_n)|\ge2m_0$. Uniform continuity gives $\delta>0$ such that
$|q|\ge m_0$ with a constant sign on $[t_n,t_n+\delta]$; for large $n$, $|r|\le m_0/2$ there. Then
$|f(t_n+\delta)-f(t_n)|\ge\delta m_0/2$, which contradicts the convergence of $f$.
\end{proof}

By Lemma~\ref{lem:wp}, $k_i\ge2\psi\sqrt b\,\hat v_{\min}>0$, and by (i), $\mu_i,\vt_i\in L_2$. All signals are bounded
by (ii) and Lemma~\ref{lem:wp}, so $\mu_i$ and $\vt_i$ have bounded derivatives and tend to zero by Barbalat's lemma. In
$\dot\vt_i=-k_i\vt_i+q_i$, the term $q_i=-b\vh_i\sinc(\vt_i)\nu_i$ is uniformly continuous, because $\vh_i$, $\vt_i$ and
$\nu_i$ have bounded derivatives almost everywhere; Lemma~\ref{lem:barb} with $f=\vt_i$ and $r=-k_i\vt_i$ gives
$q_i\to0$, and since $\vh_i\ge\hat v_{\min}$ and $\sinc\vt_i\to1$, $\nu_i\to0$. Hence $\eps_i\to0$. Under
Assumption~\ref{A4}(a), a link either has $a_{ij}(t)\to0$ or delivers samples $\eps_j(t-\tau_{ij})$ from times that tend
to infinity, so $\varpi_i\to0$ in \eqref{eq:net}.

It remains to show the following claim: for \eqref{eq:net} with measurable $\chi_i(t)\in[\chi_0,1]$, $\chi_0>0$,
bounded measurable inputs $\varpi_i$, $d_i\le d$ and links that satisfy Assumption~\ref{A4}(a), from any initial time and
with any bounded history,
\begin{equation}\label{eq:limsup}
\limsup_{t\to\infty}M(t)\le\frac1\lambda\limsup_{t\to\infty}P(t),\qquad P(t)=\max_i|\varpi_i(t)|.
\end{equation}
In the closed loop, (ii) bounds $M$ and $g_i$ and hence gives $\chi_i\ge\chi_0$, so \eqref{eq:limsup} yields
$\zeta_i\to0$. Then $g_i\to0$, $u_i\to0$, $w_i\to0$ by $\dot w_i=\gamma(u_i-w_i)$, $z_i=\zeta_i-w_i/\gamma\to0$,
$e_i=z_i+\eps_i\to0$, and $\theta_{ri}-\theta_i=\vt_i-(\thh_i-\theta_{ri})\to0$ by Lemma~\ref{lem:wp}.

To prove the claim, note that the argument of (ii) bounds $M$. Let $m$ and $p$ be the two limits superior, suppose that
$m>p/\lambda$, and let $\delta\in(0,m)$. A link either has $a_{ij}(t)\to0$, so that its delayed term is at most
$a_{ij}(t)\sup M\to0$, or has $t-\tau_{ij}(t)\to\infty$, so that its samples eventually stem from times with
$M\le m+\delta$. Hence there is a $T$ such that, for all $t\ge T$ and all $i$, $P(t)\le p+\delta$,
$\sum_ja_{ij}|\zeta_j(t-\tau_{ij})|\le d_i(m+\delta)+\delta$ and, by \eqref{eq:net},
\[
D^{+}|\zeta_i|\le\chi_i\bigl[-\lambda|\zeta_i|+\Lambda d_i(m+\delta-|\zeta_i|)+(\Lambda+1)\delta+p\bigr].
\]
Whenever $|\zeta_i|\ge m-\delta$, the bracket is at most $-\lambda m+p+(\lambda+2\Lambda d+\Lambda+1)\delta$, which is
below $-(\lambda m-p)/2$ for small $\delta$, and then $D^{+}|\zeta_i|\le-\chi_0(\lambda m-p)/2$. Every $|\zeta_i|$
therefore falls below $m-\delta$ in finite time and cannot rise above it again, which contradicts the definition of
$m$. \hfill$\blacksquare$

\subsection{Proof of Lemma~\ref{lem:strict}}

\emph{Unit scale:} Take $\ell=1$ and write $\alpha=\alpha_1$ and $U=U_1$. The quantity $B$ is at least
$2\psi\sqrt b+\sqrt b/\psi$, hence $\alpha\sqrt b\le\psi/(1+2\psi^2)$, which implies both $\alpha\sqrt b\le1/(2\sqrt2)$
and $\alpha\le\psi/\sqrt b$.

\emph{Bounds on $U$:} By the inequality of arithmetic and geometric means, $|\nu\vt|\le\sqrt b\,W$; with
$1/(1+W)\le1$ and $\alpha\sqrt b\le1/(2\sqrt2)$ this gives $c_1W\le U\le c_2W$ and the bounds on $c_1$ and $c_2$.

\emph{Derivative:} Along \eqref{eq:loop},
\begin{align*}
\dot U&=\Bigl(1+\frac{\alpha\sqrt b}{4}-\frac{\alpha\nu\vt}{(1+W)^2}\Bigr)\dot W+\frac{\alpha}{1+W}\frac{\mathrm d}{\mathrm dt}(\nu\vt),\\
\frac{\mathrm d}{\mathrm dt}(\nu\vt)&=-b\vh\sinc(\vt)\nu^2-k\nu\vt-\omega\mu\vt+\vh\vt\sin\vt .
\end{align*}
The factor of $\dot W\le0$ is at least 1, because $|\nu\vt|/(1+W)^2\le\sqrt b\,W/(1+W)^2\le\sqrt b/4$. In the second
part, write $\sinc\vt=1-(1-\sinc\vt)$, insert $\omega=\wh+k\vt+b\vh\sinc(\vt)\nu$, and use $1-\sinc\vt\le\vt^2/6$,
$\mu^2,\nu^2\le2W$, $\vt\sin\vt\le\vt^2$, $W/(1+W)\le1$ and $\sqrt{2W}/(1+W)\le1/\sqrt2$:
\begin{align*}
\tfrac{\alpha}{1+W}\,b\vh(1-\sinc\vt)\nu^2&\le\alpha\tfrac{b\vh}{3}\vt^2,\\
\tfrac{\alpha}{1+W}\,k|\nu\vt|&\le\tfrac{\alpha}{1+W}\tfrac{b\vh}{2}\nu^2+\alpha\tfrac{k^2}{2b\vh}\vt^2,\\
\tfrac{\alpha}{1+W}\,|\wh\mu\vt|&\le\alpha\tfrac{|\wh|}{2}(\mu^2+\vt^2),\\
\tfrac{\alpha}{1+W}\,k|\mu|\vt^2&\le\alpha\tfrac{k}{\sqrt2}\vt^2,\\
\tfrac{\alpha}{1+W}\,b\vh|\mu\nu\vt|&\le\alpha\tfrac{b\vh}{2\sqrt2}(\mu^2+\vt^2),\\
\tfrac{\alpha}{1+W}\,\vh\vt\sin\vt&\le\alpha\vh\vt^2 .
\end{align*}
The gain schedule gives $|\wh|\le k/(2\psi)$, $\vh\le k/(2\psi\sqrt b)$ and $k/\vh\le2\psi\sqrt{\kappa^2+b}$. With these
bounds, the collected $\mu^2$-terms are at most $\frac{\alpha k}{4\psi}(1+\sqrt{b/2})\mu^2\le\frac k2\mu^2$ by the first
entry of $\alpha$, and $2b/k$ times the collected coefficient of $\vt^2$ is at most
$\alpha[2\psi\sqrt{\kappa^2+b}+\sqrt2b+\frac b\psi(c_\alpha\sqrt b+\frac12+\frac1{\sqrt b})]=\alpha B\le1$ by the second
entry. Hence
\[
\dot U\le-\frac k2\mu^{2}-\frac{k}{2b}\vt^{2}-\frac{\alpha b\vh\,\nu^{2}}{2(1+W)}\le-\frac{\alpha b\vh\,W}{1+W},
\]
where the second inequality follows because $\alpha\le\psi/\sqrt b$ and $k\ge2\psi\sqrt b\,\vh$ imply
$\alpha b\vh\le k/2$, while $W/(1+W)\le\frac12(\mu^2+\nu^2/(1+W)+\vt^2/b)$.

\emph{General scale:} The variables $\tilde\mu=\mu/\ell$, $\tilde\nu=\nu/\ell$ and $\vt$ satisfy \eqref{eq:loop} with
$\tilde v=\vh/\ell$ and $\tilde b=b\ell^2$ in place of $\vh$ and $b$, with the same $k$ and $\omega$, and
$|\wh|\le(\kappa\ell)\tilde v$. Applying the unit-scale result to them and multiplying by $\ell^2$ gives the claims with
$\alpha_\ell=\ell\,\alpha(\psi,b\ell^2,\kappa\ell)$; note that $\alpha_\ell\sqrt b=\alpha(\psi,\tilde b,\kappa\ell)\sqrt{\tilde b}$.
\hfill$\blacksquare$

\subsection{Proof of Lemma~\ref{lem:halanay}}

The right-hand side of the equation for $\eta$ does not increase with $\eta$, equals $\chi_0(1-\beta)\lambda>0$ at
$\eta=0$, and tends to $-\infty$ if $\Lambda dh>0$ (otherwise it is constant), so the root exists and is unique. Fix
$0<\eta'<\eta$, $\delta>0$ and $T>t_0$, let $\bar P=\sup_{[t_0,T]}P$, and let
$m(t)=(1+\delta)\max\{M_h(t_0)\ee^{-\eta'(t-t_0)},\bar P/(\beta\lambda)\}$. If $m\equiv0$, the history and the input
vanish, and the solution of \eqref{eq:net} is zero by uniqueness. Otherwise $M(t_0)<m(t_0)$. Suppose that $M$ exceeds
$m$ somewhere on $[t_0,T]$, and let $t^*$ be the supremum of the times up to which $M\le m$; then $M(t^*)=m(t^*)>0$.
Since $m(s)\le m(t^*)\ee^{\eta'(t^*-s)}$ for $s\le t^*$, and also $M(s)\le M_h(t_0)\le m(t^*)\ee^{\eta'(t^*-s)}$ for
$s\in[\max\{0,t_0-h\},t_0)$, we have $M_h(t^*)\le\ee^{\eta'h}M(t^*)$; moreover,
$P\le\bar P\le\beta\lambda M(t^*)/(1+\delta)$ on $[t_0,T]$. Let $B'=(1-\beta)\lambda-\Lambda d(\ee^{\eta'h}-1)$; since
$\eta'<\eta$, $\chi_0B'\ge\eta>\eta'$. For every $i$ with $|\zeta_i(t^*)|=M(t^*)$ and almost all $t$ in a right
neighborhood of $t^*$,
\begin{align*}
D^{+}|\zeta_i|&\le\chi_i\bigl[-(\lambda+\Lambda d_i)|\zeta_i|+\Lambda d_i\,M_h(t)+P\bigr]\\
&\le-\chi_0\Bigl[B'+\beta\lambda\Bigl(1-\frac{1}{1+\delta}\Bigr)\Bigr]M(t^*)+o(1),
\end{align*}
where $o(1)\to0$ as $t\downarrow t^*$ by the continuity of $\zeta_i$ and $M_h$; because the bracket is negative,
$\chi_i$ may be replaced by $\chi_0$ and $d_i$ by $d$. The right-hand side is below $-\eta'M(t^*)$ near $t^*$, whereas
$D^{+}m\ge-\eta'm$ and $m(t)\to M(t^*)$. The other indices stay below $m$ near $t^*$. Hence $M\le m$ just after $t^*$,
a contradiction. Letting $\eta'\to\eta$ and $\delta\to0$ proves the claim. \hfill$\blacksquare$

\subsection{Proof of Theorem~\ref{thm:exp}}

By Lemma~\ref{lem:wp}, Lemma~\ref{lem:strict} applies with $\kappa=\bar\kappa$ and $\vh_i\ge\hat v_{\min}$. Since
$W\mapsto W/(1+W/\ell^2)$ increases and $W\ge U_\ell/c_2$, Lemma~\ref{lem:strict} gives
$\dot U_\ell\le-\alpha_\ell b\hat v_{\min}U_\ell/(c_2+U_\ell/\ell^2)$, that is,
$\frac{\mathrm d}{\mathrm dt}(c_2\ln U_\ell+U_\ell/\ell^2)\le-\alpha_\ell b\hat v_{\min}$. Integrating, dropping
$U_\ell(t)/\ell^2\ge0$ and using $c_1W\le U_\ell\le c_2W$ gives
$W_i(t)\le(c_2/c_1)\,\ee^{W_i(0)/\ell^2}W_i(0)\,\ee^{-2\eta_1t}$. With $W_i(0)\le E_0^2/2$ and $|\eps_i|\le\sqrt{2W_i}$,
this is the first estimate, and $|\vt_i|\le\sqrt{2bW_i}\le\sqrt b\,C_1E_0\ee^{-\eta_1t}$.

Let $\Lambda d>0$ (otherwise the filters stay at rest). Since $\eta_2>0$ solves
$\eta_2=\chi_0[\lambda/2-\Lambda d(\ee^{\eta_2h}-1)]$, we have $\ee^{\eta_2h}<1+\lambda/(2\Lambda d)$, and
$2\eta^*\le\min\{\eta_1,\eta_2\}$. A delayed sample stems from a time $s\ge\max\{0,t-h\}$, at which
$|\eps_j(s)|\le C_1E_0\ee^{-2\eta^*\max\{0,t-h\}}\le C_1E_0\ee^{\eta_2h}\ee^{-2\eta^*t}$. With \eqref{eq:net}, this gives
\begin{align*}
P(t)&\le\Lambda dC_1E_0\bigl(1+\ee^{\eta_2h}\bigr)\ee^{-2\eta^*t}\\
&\le\Bigl(2\Lambda d+\frac\lambda2\Bigr)C_1E_0\,\ee^{-2\eta^*t}.
\end{align*}
Lemma~\ref{lem:halanay} on $[t/2,t]$ with $\beta=\tfrac12$, $M_h\le2\Lambda dE_0/\lambda\le C_2E_0$, where
$C_2=(1+4\Lambda d/\lambda)C_1$, and $\chi_i\ge\chi_0$ from Theorem~\ref{thm:conv}(ii) gives
$M(t)\le\max\{C_2E_0\ee^{-\eta_2t/2},(2/\lambda)\sup_{[t/2,t]}P\}\le C_2E_0\ee^{-\eta^*t}$. A delayed $\zeta_j$ then
satisfies $|\zeta_j(s)|\le C_2E_0\ee^{\eta^*h}\ee^{-\eta^*t}$, and
$\Lambda d\,\ee^{\eta^*h}\le\Lambda d\,\ee^{\eta_2h/2}\le\Lambda d\sqrt{1+\lambda/(2\Lambda d)}\le\Lambda d+\lambda/4$.
Hence, with $C_3=(\tfrac54\lambda+2\Lambda d)C_2+(2\Lambda d+\tfrac12\lambda)C_1$,
\begin{align*}
|u_i|\le|g_i|&\le(\lambda+\Lambda d)M(t)+\Lambda d\,C_2E_0\ee^{\eta^*h}\ee^{-\eta^*t}+P(t)\\
&\le C_3E_0\ee^{-\eta^*t},
\end{align*}
and $\dot w_i=\gamma(u_i-w_i)$ with $w_i(0)=0$ and $\eta^*\le\gamma/2$ gives $|w_i|\le2C_3E_0\ee^{-\eta^*t}$. From
$z_i=\zeta_i-w_i/\gamma$, $e_i=z_i+\eps_i$, $|\thh_i-\theta_{ri}|\le\tfrac\pi2|w_i|/v_{\min}$ (Lemma~\ref{lem:wp}) and the
bound on $\vt_i$, the claims follow with
\[
K=C_1+C_2+\frac{2C_3}{\gamma},\qquad K'=\sqrt b\,C_1+\frac{\pi C_3}{v_{\min}} .
\]
None of $C_1$, $C_2$, $C_3$, $K$, $K'$ depends on $h$. \hfill$\blacksquare$

\subsection{Asymptotic Rate (Remark~\ref{rem:uas})}

Under Assumption~\ref{A4}(b), Theorem~\ref{thm:conv}(iii) and its proof give $\zeta_i,w_i,u_i,g_i\to0$. Hence
$\vvh_i-v_{ri}\xi(\theta_{ri})\to0$ and $\avh_i-\dot v_{ri}\xi(\theta_{ri})-v_{ri}\omega_{ri}\xip(\theta_{ri})\to0$, and
since $v_{ri}\ge v_{\min}>0$, $\vh_i-v_{ri}\to0$ and $\wh_i-\omega_{ri}\to0$. For every $\delta\in(0,v_{\min})$ there is
therefore a $T_\delta$, which depends on the solution, such that $|\wh_i|\le(\kappa_r+\delta)\vh_i$ and
$\vh_i\ge v_{\min}-\delta$ for all $i$ and $t\ge T_\delta$. Lemma~\ref{lem:strict} with $\kappa=\kappa_r+\delta$ and the
integration in the proof of Theorem~\ref{thm:exp}, started at $T_\delta$, give
$|\eps_i(t)|\le c\,\ee^{-\eta_\delta(t-T_\delta)}$ with $\eta_\delta=\alpha_\ell b(v_{\min}-\delta)/(2c_2)$ and
$\alpha_\ell,c_2$ evaluated for $\kappa=\kappa_r+\delta$. As $\delta\to0$, $\eta_\delta$ tends to the rate
$\eta_1^{\infty}$ stated in Remark~\ref{rem:uas}, by the continuity of $\alpha$ in $\kappa$. \hfill$\blacksquare$

\subsection{Local Input-to-State Estimate (Remark~\ref{rem:uas})}

\begin{corollary}\label{cor:iss}
Let \eqref{eq:model} be perturbed to $\dot p_i=(v_i+\delta_{vi})\xi(\theta_i)+\delta_{ni}\xip(\theta_i)$,
$\dot\theta_i=\omega_i+\delta_{\omega i}$ with measurable, locally bounded perturbations, let Assumptions
\ref{A1}--\ref{A3} and \ref{A5} hold and the filters start at rest, fix $\ell>0$, and let
$\Delta_i=\frac54|\delta_{vi}|+\frac74|\delta_{ni}|+(\frac{7}{4\sqrt b}+0.18\ell)|\delta_{\omega i}|$. For every
$W_{\max}>0$ there are $\Delta_{\max},\eta_3,G>0$, given below and independent of the delays and the graph, such that if
$U_\ell(0)\le c_1W_{\max}$ for every vehicle and $\Delta_i(t)\le\Delta_{\max}$ for all $i$ and $t$, then the solution
exists on $[0,\infty)$, $W_i\le W_{\max}$,
\[
|\eps_i(t)|\le\sqrt{2c_2W_i(0)/c_1}\;\ee^{-\eta_3t/2}+G\sup_{s\le t}\Delta_i(s),
\]
and $|e_i(t)|\le C[\sqrt{2c_2W_0/c_1}+G\sup_{j,s}\Delta_j(s)]$ with $W_0=\max_jW_j(0)$ and $C$ of
Theorem~\ref{thm:conv}.
\end{corollary}

\begin{proof}
With the perturbations, $\dot W=-k\mu^2-(k/b)\vt^2-\mu\delta_v-\nu\delta_n-\vt\delta_\omega/b$ (index $i$ dropped), and
$\frac{\mathrm d}{\mathrm dt}(\nu\vt)$ acquires the terms $-\delta_\omega\mu\vt-\delta_n\vt-\delta_\omega\nu$. The bounds
of Lemma~\ref{lem:wp} use only $|u_i|\le\rho v_{ri}$ and remain valid; local existence follows as in its proof, and
the invariance shown below excludes a finite escape. The factor of $\dot W$ in $\dot U_\ell$ lies in
$[1,\tfrac54]$, since $\alpha_\ell\sqrt b\le\tfrac12$. With $|\mu|,|\nu|\le\sqrt{2W}$, $|\vt|\le\sqrt{2bW}$,
$|\mu\vt|\le\sqrt bW$, $\sqrt W/(1+W/\ell^2)\le\ell/2$ and $\alpha_\ell\sqrt b\le1/(2\sqrt2)$, the new terms are bounded
by $\sqrt{2W}\,\Delta$, so that, with $\varsigma=\alpha_\ell b\hat v_{\min}$, for all states and almost all $t$
\[
\dot U_\ell\le F(W):=-\frac{\varsigma W}{1+W/\ell^2}+\sqrt{2W}\,\Delta_{\max} .
\]
\emph{Constants:} $\Delta_{\max}=\varsigma(1+W_{\max}/\ell^2)^{-1}\sqrt{c_1W_{\max}/(2c_2)}$,
$\eta_3=\varsigma/(2c_2(1+W_{\max}/\ell^2))$ and $G=2(1+W_{\max}/\ell^2)\varsigma^{-1}\sqrt{c_2/c_1}$.

\emph{Invariance:} Let $\mathcal U=c_1W_{\max}$. Since $\sqrt W/(1+W/\ell^2)$ is unimodal, it is at least
$\sqrt{c_1W_{\max}/c_2}/(1+W_{\max}/\ell^2)$ on $[c_1W_{\max}/c_2,W_{\max}]$, and $F\le0$ there by the definition of
$\Delta_{\max}$. Let $L_F$ be a Lipschitz constant of $F$ on $[c_1W_{\max}/c_2,2W_{\max}]$. Where
$U_\ell\in[\mathcal U,2\mathcal U]$, we have $W\in[U_\ell/c_2,U_\ell/c_1]\subset[c_1W_{\max}/c_2,2W_{\max}]$ and hence
$F(W)\le L_F(W-W_{\max})_+\le(L_F/c_1)(U_\ell-\mathcal U)$. Suppose that $U_\ell(t_1)>\mathcal U$ for some $t_1$, and let
$t_0<t_1$ be the last time before $t_1$ with $U_\ell(t_0)=\mathcal U$; after moving $t_1$ closer to $t_0$ if necessary,
$\mathcal U<U_\ell\le2\mathcal U$ on $(t_0,t_1]$. Then $y=U_\ell-\mathcal U$ is absolutely continuous, $y(t_0)=0$ and
$\dot y\le(L_F/c_1)y$ almost everywhere on $[t_0,t_1]$, so that $y\equiv0$ on $[t_0,t_1]$ by Gronwall's inequality, a
contradiction. Hence $\{U_\ell\le c_1W_{\max}\}\subset\{W\le W_{\max}\}$ is forward invariant.

\emph{Estimates:} On this set, $\sqrt{2W}\,\Delta\le\frac{\varsigma W}{2(1+W_{\max}/\ell^2)}+\frac{(1+W_{\max}/\ell^2)\Delta^2}{\varsigma}$
by Young's inequality, so that $\dot U_\ell\le-\eta_3U_\ell+(1+W_{\max}/\ell^2)\Delta^2/\varsigma$, and the comparison
lemma gives $U_\ell(t)\le U_\ell(0)\ee^{-\eta_3t}+\frac{2c_2(1+W_{\max}/\ell^2)^2}{\varsigma^2}\sup_{s\le t}\Delta^2$; with
$|\eps|\le\sqrt{2U_\ell/c_1}$ and $U_\ell(0)\le c_2W(0)$, taking square roots gives the estimate of $\eps_i$. The argument
of Theorem~\ref{thm:conv}(ii) uses only the bound on $\eps$, not the monotonicity of $W_i$; with $E_0$ replaced by
$\sup_{j,s}|\eps_j(s)|\le\sqrt{2c_2W_0/c_1}+G\sup\Delta$, it gives the bound on $e_i$.
\end{proof}

\subsection{Derivation of Remark~\ref{rem:freq}}

Below $\bar e$ and with $\chi_i=1$, \eqref{eq:net} reads
$\dot\zeta=-[(\lambda I+\Lambda D)\otimes I_2]\zeta+(\Lambda A\otimes I_2)\zeta(t-\tau)-(\Lambda D\otimes I_2)\eps
+(\Lambda A\otimes I_2)\eps(t-\tau)$ for constant weights and a uniform delay. Since $\dot\zeta_i=u_i$ and
$\dot w_i=\gamma(u_i-w_i)$, the Laplace transforms with zero initial data satisfy
$z_i=\zeta_i-w_i/\gamma=\zeta_i/(1+s/\gamma)$, and $e_i=z_i+\eps_i$ gives \eqref{eq:H}. The poles of $H$ are $-\gamma$ and
the roots of $\det(sI+\lambda I+\Lambda D-\Lambda A\ee^{-s\tau})=0$; by Lemma~\ref{lem:halanay} with $\chi_0=1$ and
$P=0$, whose proof applies verbatim to a constant delay with an initial history on $[-\tau,0]$, every solution of the
unforced network decays exponentially, so these roots lie in the open left half-plane. On
the complete graph of three vehicles, $D=2I$ and $A=-I$ on the range of $\Pi$, so that $H$ acts there as the scalar
\[
h_3(s)=1-\frac{\Lambda(2+\ee^{-s\tau})}{(1+s/\gamma)(s+\lambda+2\Lambda+\Lambda\ee^{-s\tau})} .
\]
\hfill$\blacksquare$

\subsection{Proof of Proposition~\ref{prop:direct}}

On a straight reference along the first axis with constant speed $v$, $\omega_{ri}=0$ and $k_i=2\psi\sqrt b\,v=:k$.
Linearizing at $e_i=0$, $\vt_i=0$ with $s_i=(s_{i\parallel},s_{i\perp})$ gives $\dot e_{i\parallel}=-ks_{i\parallel}$,
$\dot e_{i\perp}=v\vt_i$ and $\dot\vt_i=-k\vt_i-bv\,s_{i\perp}$, so that
$\ddot e_{i\perp}=-k\dot e_{i\perp}-bv^2s_{i\perp}$, and stacking with $s_\perp=(I+cD)e_\perp-cAe_\perp(t-\tau)$,
$D=dI$, gives \eqref{eq:lin}. The eigenvectors of the symmetric matrix $A$, with eigenvalues $a\in[-d,d]$, decouple
the modes $s+k(1+cd)-kca\,\ee^{-s\tau}$ (longitudinal), which are stable for every $\tau$ because $1+cd>c|a|$, and
$p(s)+q\,\ee^{-s\tau}$ with $p(s)=s^2+ks+bv^2(1+cd)$ and $q=-bv^2ca$ (lateral). For $\tau=0$ these are stable, since
$1+c(d-a)>0$, and $p$ is Hurwitz. By the frequency-sweeping test \cite{gu2003}, a lateral mode is stable for every
$\tau\ge0$ if and only if $|p(j\omega)|>|q|$ for all $\omega\ge0$. With $\omega=\sqrt b\,v\,\Omega$ and $x=cd$,
$|p(j\omega)|^2/(bv^2)^2=(1+x-\Omega^2)^2+4\psi^2\Omega^2$, whose minimum over $\Omega$ is $(1+x)^2$ if
$1+x\le2\psi^2$ and $4\psi^2(1+x)-4\psi^4$ otherwise. Since $|q|\le bv^2x$ with equality for $a=d$, all modes are stable
for every delay if and only if this minimum exceeds $x^2$, that is, if and only if $x<2\psi(1+\psi)$. Otherwise, the
mode $a=d$ (eigenvector $\mathbf1$) has a root $j\omega_c$ at the delays $\tau=(\theta_c+2\pi m)/\omega_c$,
$m=0,1,\dots$, where $\omega_c>0$ solves $|p(j\omega_c)|=|q|$ and $\theta_c\in[0,2\pi)$ is the argument of
$-q/p(j\omega_c)$; in the variables $(\Omega,\sqrt b\,v\tau)$ these equations depend only on $\psi$ and $x$, so the
smallest such delay is proportional to $1/v$. Roots cross into the right half-plane at the larger crossing frequency and back at the smaller
one, which gives the region of Fig.~\ref{fig:kin}(a). \hfill$\blacksquare$

\fi

\section*{ACKNOWLEDGMENT}

The author used the large language model Claude \cite{claude} (Anthropic) to assist with writing the simulation code,
and reviewed the code and all results.

%%%%%%%%%%%%%%%%%%%%%%%%%%%%%%%%%%%%%%%%%%%%%%%%%%%%%%%%%%%%%%%%%%%%%%%%%%%%%%%%

\end{document}